\RequirePackage{fix-cm}
\documentclass[twocolumn]{svjour3}          % twocolumn
\smartqed  % flush right qed marks, e.g. at end of proof
\usepackage{graphicx}
\usepackage[square,sort,comma,numbers]{natbib}  % DO NOT CHANGE THIS AND DO NOT ADD ANY OPTIONS TO IT
\usepackage{graphicx}
\usepackage{subcaption}
\usepackage[switch]{lineno}  %
\usepackage[export]{adjustbox}
\usepackage{amsmath, amssymb}
\usepackage{dsfont}
\usepackage{comment}
\usepackage{multicol}
\usepackage{multirow}
\usepackage{booktabs}
\usepackage{hyperref}

\begin{document}

\title{Relation Transformer Network%\thanks{Grants or other notes
}
% \subtitle{A cool subtitle can go here!}


\author{Rajat Koner         \and Suprosanna Shit \and 
        Volker Tresp %etc.
}


\institute{R. Koner \at
              Ludwig Maximilian University of Munich, Munich, Germany \\
              \email{koner@dbs.ifi.lmu.de}           %  \\
           \and
           S. Shit \at
              Technical University of Munich, Munich,Germany\\
              \email{suprosanna.shit@tum.de}
          \and
          V. Tresp \at 
          Siemens AG, Munich, Germany\\
          \email{volker.tresp@siemens.com}
}

\date{Preprint : under review}

\maketitle

\begin{abstract}
The extraction of a scene graph with objects as nodes and mutual relationships as edges is the basis for a deep understanding of image content. Despite recent advances, such as message passing and joint classification, the detection of visual relationships remains a challenging task due to sub-optimal exploration of the mutual interaction among the visual objects. In this work, we propose a novel transformer formulation for scene graph generation and relation prediction. We leverage the encoder-decoder architecture of the transformer for rich feature embedding of nodes and edges. Specifically, we model the node-to-node interaction with the self-attention of the transformer encoder and the edge-to-node interaction with the cross-attention of the transformer decoder. Further, we introduce a novel positional embedding suitable to handle edges in the decoder. Finally, our relation prediction module classifies the directed relation from the learned node and edge embedding. We name this architecture as Relation Transformer Network (RTN). On the Visual Genome and GQA dataset, we have achieved an overall mean of {4.85\% and 3.1\%} point improvement in comparison with state-of-the-art methods. Our experiments show that Relation Transformer can efficiently model context across various datasets with small, medium, and large-scale relation classification.

% , which preserves the local properties of an edge to generate a  $\langle subject, predicate, object\rangle$ triple.

% We have conducted several experiments on the Visual Genome, GQA, and the VRD dataset to
% demonstrate the efficiency of Relation Transformer's context accumulation.

% , which captures dependencies between objects and
% edges and their effects on relational labels.
\end{abstract}

%%%%%%%%%%%%%%%%%%%%%%%%%%%%%%%%%%%%%%%%%
\section{Introduction}
\label{sec:intro}
\begin{figure}[t]

\begin{subfigure}{0.5\textwidth}
\centering
\includegraphics[width=0.85\linewidth, height=5.5cm]{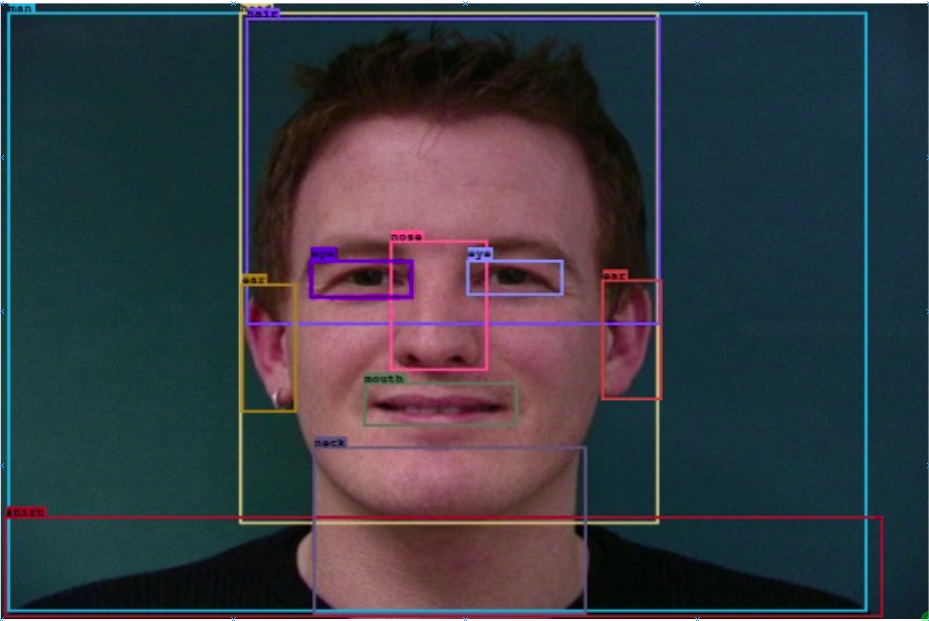} 
\caption{Scene consisting of a man's  face}
\label{fig:subim1}
\end{subfigure}
\begin{subfigure}{0.5\textwidth}
\centering
\includegraphics[width=0.85\linewidth, height=5.5cm]{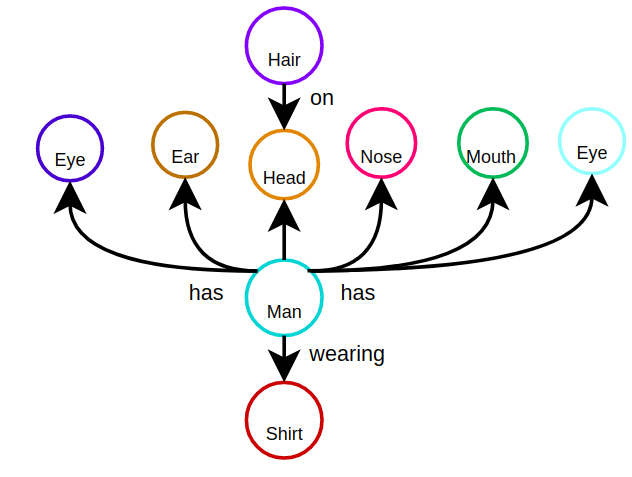}
\caption{Corresponding scene graph}
\label{fig:subim2}
\end{subfigure}
\caption{\textbf{(a)} is an example image  of a face of a man. \textbf{(b)} describes the corresponding  scene graph,  annotated with various object labels like head, ear, shirt (color coded as the respective bounding box) and their mutual relationships.}
\label{fig:man_sg}
\end{figure}

A \textit{scene graph} is a graphical representation of an image consisting of multiple entities and their relationships, expressed in a triplet format like $\langle$\textit{subject, predicate, object}$\rangle$. Objects in the scene become \textit{nodes}, undirected interactions between nodes are represented by \textit{edges} and a directed edge is called a relationship or predicate. E.g. in  Fig. \ref{fig:man_sg}, \lq Eye\rq,\lq Hair\rq,\lq Head\rq, and \lq Man\rq\  are object or node labels and their mutual relationships are described by the predicates \lq has\rq\  and \lq on\rq. An  extracted scene graph can be used in many downstream applications like visual question answering \cite{ghosh2019generating,hildebrandt2020scene,koner2021graphhopper}, image retrieval \cite{Schroeder2020}, and  image captioning \cite{xu2019scene}.
 
A scene graph generation (SGG) task is executed in  two steps: first, objects present in the image are detected, and second, the most suitable predicates are determined for selected object pairs. Current object detection approaches have achieved outstanding performance in spatially locating objects in an image. In contrast,  performance on relation prediction is still quite limited. Several recent works have tried to explore SGG from different perspectives. Context information  is exchanged either globally \cite{zellers2018neural} or across neighborhoods \cite{Xu_2017_CVPR,yang2018graph}.  \cite{newell2017pixels, zhang2019graphical}  introduced a variant of contrastive loss for a better representation of similar types of relations and some contemporary work \cite{lin2020gps} improved message passing with direction and priority sensitive loss. \cite{knyazev2020graph} addresses specifically  the biases in different datasets. 
 Interactions among objects and their corresponding edges through contextual attention for SGG, the topic of this paper,  is still under-explored.
 % We argue in this paper, an efficient propagation of scene context information for both nodes and edges could provide a state of the art results without any rings and bells.
 This paper proposes a novel transformer-based formulation for the SGG task, namely Relation Transformer network (RTN). In the following, we rationalize the self-attention and cross-attention of the proposed RTN in context of SGG task.
 
First, it is crucial to understand the role of each object in an image and how object labels are related and influenced by others in the context of the whole image. 
For example,  in Fig. \ref{fig:man_sg}, the presence of node labels like \lq Eye\rq, \lq Hair\rq, \lq Nose\rq, \lq Ear\rq\ indicate that these together describe a face and indicate the presence of a node with label  \lq Face\rq\ or \lq Head\rq\ in their surroundings.    Additionally, the node label \lq Shirt\rq\ implies that this is a face or head of a \lq Human\rq\ and not an animal. The spatio-semantic co-occurrence of labels along with the contextualization of nodes are both important for predicting pairwise relations and node labels. To obtain ``context aware nodes'', we have modeled interactions among nodes using self-attention of the transformer \cite{vaswani2017attention}, which in this case is a node-to-node (N2N) attention propagation.
 
Second, a subsequent challenge is to predict the exact relationship between two objects. In this paper, we assume that node labels and edge labels are mutually dependent and predicted by scene context. %if it is aware of all the nodes or vice versa as well as its source node($\langle subject, object \rangle$). This unidirectional edge could also effectively capture a directed relation for joint classification($\langle subject, edge,object \rangle$) of a predicate.
For example, in Figure \ref{fig:man_sg}, the probabilities for 
edges between the node \lq Man\rq\ and its body parts like \lq Head\rq\ ,\lq Eye\rq\ and \lq Nose\rq\ should be similar because of the spatio-semantic similarity.
Thus context propagation from nodes to edges directly influences the classification of their directed relations ($\langle Man,~has,~Head\rangle,\langle Head,~of,~Man\rangle$). Although a directed edge label is mainly dependent on its two associated nodes, i.e., \textit{subject} and \textit{object}, we argue that attention from all other nodes to an edge helps to leverage all mutually correlated relations and guides towards identifying most consistent relationship in the global scene context. This edge to nodes interaction is modeled using the cross-attention of the transformer decoder, which we refer to as E2N attention.

% because of their global view. %as each of this node and edge influence each other according the scene in a multi-hop manner. 

 % 
%  for an efficient global context exploration the edge should exploit this dependency while exploring all nodes.} 
In addition, we introduce a novel positional encoding that preserves the local context of the edge. Finally, a directed relation prediction module (RPM) \ref{sub:rpm} is used to classify the relation that takes advantage of the context-rich nodes and edges. To summarize our contributions:

\begin{enumerate}
\setlength\itemsep{0pt}
\item We formulate the SGG task as a feature learning on a graph leveraging transformer architecture to model object-to-object and object-to-relation interaction. First, we extract object-to-object interaction via the self-attention (N2N attention) of the transformer encoder. 
\item Next, we use this rich node representation to capture object-to-relation interaction via the transformer decoder's cross-attention (E2N attention). We introduce a novel positional encoding for the edges in the transformer decoder to accumulate a global scene environment while preserving local context.
\item We employ an efficient directional relation prediction module to accumulate the learned node and edge representation from the transformer and classify the desired directed relation.
\item We perform extensive experiments in multiple datasets to show our proposed RTN's generalizability and efficacy. We achieve an overall mean of 4.85\% and 3.2\% improvement on the challenging Visual Genome and GQA dataset over the state-of-the-art models.
\end{enumerate}

\section{Related Work}
\textbf{Scene Understanding with Language Prior:}
Scene understanding evolved through many phases throughout recent years. Initially, researchers tried to localize objects or regions in an image, based on a given caption or text reference \cite{mao2016generation,nagaraja2016modeling,hu2017modeling,plummer2015flickr30k}. These approaches mostly matched the referenced text to the matching part of the image. \cite{johnson2015image} introduced scene graphs for image retrieval, and \cite{lu2016visual}, proposed visual relationship detection with language priors, and introduced an associated dataset named VRD. \cite{baier2017improving} derived a knowledge graph model from the training data labels and achieved generalization to new triples by knowledge graph factorization approaches. Several works focused on combining the visual and other semantic features of the subject, object, and predicates \cite{zhang2019graphical,newell2017pixels,wan2018representation,yin2018zoom,zhang2017visual,lu2016visual} and enriched the features with a variant of triplet loss, pooling, and multi-modal representations \cite{sah2019improving}.

\noindent\textbf{Context in Scene Graph:} Contextual information has been shown to be helpful for object detection \cite{liu2018structure}, visual question answering \cite{Anderson_2018_CVPR}, and  scene understanding \cite{nagaraja2016modeling}. Recent advancements in the attention approach provided an efficient way to model complex interactions of entities in NLP networks \cite{vaswani2017attention}, and convolution networks \cite{wang2018non}.  %\textit{Also, the evolution of the graph structure network especially graph convolution \cite{kipf2016semi,velivckovic2017graph} broadens the research.} 
Various recent relationship detection networks have  tried to incorporate context with attention or transformer \cite{zareian2020learning,woo2018linknet,yang2018graph,lin2020gps} or without attention \cite{qi2019attentive,zellers2018neural,herzig2018mapping,Xu_2017_CVPR}. Although our work is also based on attentional context, it differs as it introduces the novel  N2N and E2N attention. Furthermore,  like \cite{zellers2018neural}, we do not only consider interactions between mutually co-occurrence objects, but we also analyze how the presence of objects or predicates jointly influences each other.

\noindent\textbf{Transformers in Vision:}
After the release of the transformer \cite{vaswani2017attention}, 
it became one of the most popular approaches for various vision \cite{koner2021oodformer,carion2020end} or vision language tasks from natural language processing \cite{li2019unicoder}. In Vision-Language pretraining tasks, BERT-style architectures \cite{devlin2018bert}  became a default choice, due to their ability to process sequential and also non-sequential data and, in almost all cases,  it improved upon the state of the art results. In \cite{lu2019vilbert}, a two-stream network for joint vision-language modalities has been used to obtain an enhanced representation for tasks like visual question answering and image captioning. \cite{li2019unicoder}  uses a combination of sentences and image patches jointly for pretraining and achieved a state of the result on GQA \cite{hudson2019gqa} or tasks like Masked Object Classification (MOC), Visual Linguistic Matching (VLM). Recently,  \cite{carion2020end} proposed a simple end-to-end object-detection framework; similar to our work,  it uses a  transformer encoder-decoder architecture. This recent surge of interest shows the importance and efficacy of the transformer and BERT-style architectures. In comparison to similar approaches, the  Relation Transformer in this paper gives superior performance and interpretable results; at the same time, it has a transparent modular architecture. An earlier study of our method can be found at \cite{koner2021scenes}. 

% provides a unique advantage of modular, simple yet effective frame work for node and edge based image contextualization.
\section{Method}
\label{sec:our_method}

\begin{figure*}
\includegraphics[width=0.95\textwidth]{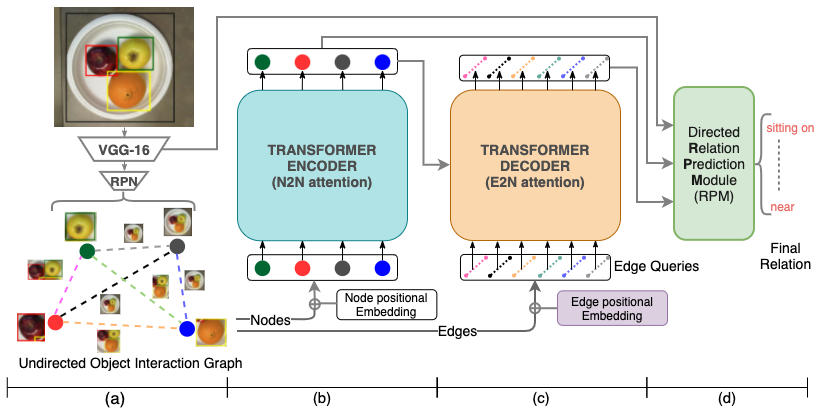}
 \caption{An overview of the proposed Relation Transformer architecture. The network consists of four stages: 
 a) Extraction of initial node (colored circle) and edge (dotted line) features using a region proposal network (RPN) of an object detector, 
 b) Creation of context-rich node embeddings using N2N attention from initial 
 node embeddings,  
 c) Creation of edge embedding by accumulating context from nodes  through E2N attention and proposed edge position embedding, 
 d) direction aware classification of the relation using $\langle subject, edge, objects \rangle$ triplet embedding. 
 Best viewed in color.}
\label{fig:atten4rel}

\end{figure*}

We frame the SGG task as a multi-hop attention-based context-propagation problem among nodes, edges, and their joint classification in a directed graph. This task is decomposed into four sub-tasks, (1) Object detection to get nodes (object bounding boxes) and their edges (union of object bounding boxes) [in Sec. \ref{od}].
(2) Modeling of interactions between the nodes [in Sec. \ref{cpo}]
(3) Accumulation of necessary context from all nodes for an edge [in Sec. \ref{sub:cee}] 
and,
(4) Classification of directed relations between the objects from the extracted context information. In the next sub-sections, we will describe these sub-tasks along with a brief introduction of the attention mechanism of the transformer. An overview of the proposed Relation Transformer architecture is shown in Fig. \ref{fig:atten4rel} with nodes and edges derived from a representative image.
% and the directed relation prediction module (RPM) [in Sec. \ref{sub:rpm}].

\subsection{Problem Decomposition}

A scene graph $G$ consists of a set of nodes $N = \{n_i\}$, that represent the objects in an image
and a set of labeled directed edges $R \subset N \times \mathcal R \times N$, where 
$\mathcal R$ is the set the relation types. For each node $n_i$, $b_i \in \mathbb{R}^4$ denotes the bounding box coordinates and $o_i$ denotes the class label. We denote $I$ as the image and  $B$ as the set of bounding boxes, and $O$ as the set of object class labels. With the help of this notation, a generative model for the graph is $Pr(B \rvert I)\: Pr(O \rvert B,I)\: Pr(R \rvert O,B,I).$ $Pr(B \rvert I)$ is inferred by our object detection module. From the objects and bounding boxes we construct an \textit{undirected object interaction graph} [Fig. \ref{fig:atten4rel}] with all the detected objects as a nodes $\{n_i\}$ and possible node-pairs with edges $\{e_{ij}\}$. Such an undirected edge is a candidate directed relation without immediate knowledge of the direction (subject to object) and interaction label (predicate). For the sake of this paper, we assume that, if $e_{ij}$ exists, there is a unique relation type. We propose to estimate $Pr(O \rvert B,I)$ by the N2N module, where other labels can influence the presence of one object label. Subsequently, to model the relationships $Pr(R \rvert O,B,I)$, we first process the candidate undirected edges $\{e_{ij}\}$ between $n_i$  and $n_j$ using the E2N module and learn an object interaction graph. Then the RPM predicts edge direction and relation type $r_{{i}\rightarrow {j}}$. 
% In this section we provide a brief introduction to KGs in
% a formal setting and review the most relevant related work.
% Let E denote the set of entities and consider the set of binary relations R. A knowledge graph KG ⊂ E × R × E i

% A scene graph $G= (N,R)$ of an image $I$ consists of
% each node or object ($n_i\in N$) and relation $r_{{i}\rightarrow {j}}\in R$ among $i^{th}$  and % $j^{th}$ node present in the image. 

% A set of nodes $\{n_i\}$, can be represented by their corresponding bounding box location as B = % \{$b_1,b_2,...,b_n$\}, $b_i\in \mathbb{R}^4$ 
%  and their class labels O = \{$o_1,o_2,...,o_n$\}, $o_i\in C$. %In this paper each relation 
% $r_{{i}\rightarrow {j}}$ defines the relationship between the subject and object node. 

\subsection{Object Detection}
\label{od}
We have used Faster-RCNN  \cite{ren2015faster} with a VGG-16  \cite{simonyan2014very} backbone for object detection.
For a node $n_i$, we obtain the \textit{spatial embedding} bounding box coordinates $b_i \in \mathbb{R}^4$, a visual feature vector of a region of interest $v_i \in \mathbb{R}^{4096}$ from  the feature map $I_{feature}$ obtained from top layer of VGG-16. Also, we get initial class probabilities $ o^{\textit{init}}_i \in \mathbb{R}^{C}$ where $C$ is the number of classes. To exploit the semantic information of the predicted class label $(o^{\textit{init}}_i)$, we multiply it with the GloVe embedding \cite{pennington2014glove} of all classes to obtain the semantic features $s_i$. This enforces hard attention of detected class probabilities across the word embedding feature space.

% where $[v^{\textit{RoI}}_i; o^{\textit{init}}_i; b_i]$\footnote{in our implementation, we have used normalized bounding boxes both for nodes and edges, please see the spatial embedding section.}  denotes the concatenation of the three vectors. 

\subsection{Context Propagation via Transformer}
\label{cp}
The core concept of our approach is the efficient
attention-based context propagation across all nodes and edges using  an encoder-decoder architecture implemented as  transformers \cite{vaswani2017attention}.
% \subsubsection{Attention:}\label{at}
The transformer architecture uses self-attention mechanisms for mapping of the global dependencies. One defines attention as the matrix
\begin{align}
    \text{Attention}(Q,K,V) = \text{softmax}(\dfrac{QK^T}{\sqrt{d_k}})V. \label{attention}
\end{align}
where query (Q), keys (K), and value (V) are obtained through three learnable layers, and $d_k$ is a scaling factor. The output is computed as a weighted sum of the values, where the weight is computed by multiplying a query matrix with its corresponding key.

In our transformer architecture, we reason two different attention schemes based on our observation as discussed in the Sec. \ref{sec:intro}. We incorporate self-attention module in the encoder of our transformer that serves as a N2N attention. However, to model the optimal contextualization from all nodes to edges, we employ E2N attention as the cross-attention in the decoder of our transformer. To exploit both global and local context propagation in the E2N attention, we introduce appropriate changes in the positional encoding of the decoder.

% We have used encoder self-attention as N2N and decoder-encoder cross-attention used as E2N  for our object and edge context propagation module.

\subsubsection{Encoder N2N Attention}
\label{cpo}
Contextualization of objects by exploring its surroundings not only enhances object detection \cite{liu2018structure}, but also encodes more discriminate features for relation classification. For this purpose, we make a permutation invariant sequential ordering of the nodes and pass this node sequence to the transformer encoder. The initial node feature vector ($f_i^{\textit{in}}$) for the  $i^{th}$ node 
is obtained by applying a linear projection layer ($W_{\textit{node}}$) on its concatenated features as
% We have considered these individual proposals and their respective features as the initial node embeddings ($n^{\textit{in}}_i$) of the scene graph.
\begin{equation}\label{nin}
f^{\textit{in}}_i = W_{\textit{node}}([v_i; s_i; b_i])
\end{equation} 
Additionally for $i^{th}$ node, we added a positional feature vector ($\textit{pos}(n_i)$) with its initial feature $f^{\textit{in}}_i$. It takes the categorical position of $i^{th}$ node in a linear ordering of all nodes, and covert it into a continuous sinusoidal vector as described in \cite{vaswani2017attention}.
\begin{align}\label{nfinal}
f^{\textit{final}}_i &= \textit{encoder}(f^{\textit{in}}_i+\textit{pos}(n_i))\\
o^{\textit{final}}_i &= \text{argmax}(W_{\textit{classifier}}(f^{\textit{final}}_i)).\label{ofinal}
\end{align}
where \textit{encoder} is a stack of multi-head attention layers as shown in Figure \ref{fig:atten4rel}. After the contextualization of the nodes by the encoder, we obtain final node features $f^{\textit{final}}_i$. This semantically enriched node feature is subsequently used for two purposes. First,  
it is passed through a linear object classifier $W_{\textit{classifier}}$ to get accurate final object class ($o^{\textit{final}}_i \in C$) probability as described in Eq. \ref{ofinal} and, second, the $f^{\textit{final}}_i$ is passed to the the decoder cross-attention (E2N attention) for edge context propagation.

% based on the actual position of the node in a linear ordering of all nodes

\subsubsection{Decoder Edge Positional Encoding}
\label{sub:pos_en}
We feed the edges of the undirected object interaction graph to the transformer decoder along with its positional embedding, which we refer to as \textit{Edge Queries}. Since there is no explicit ordering among edges we proposed a novel positional embedding for edge from both of its node position. The new positional encoding vector ($\textit{pos}_{\textit{e}_{ij}}\in \mathbb{R}^{2048}$) for edges ($e_{ij}$),  encodes the position of both the source nodes in an interleaved manner. One of these nodes  will play the roles of either subject or object. Since our edge is undirected, we hypotheses that our proposed edge positional embedding will be helpful for the  network to distinguish the source nodes (subject or object) out of all distinct nodes. The goal is to  accumulate the necessary global context (all distinct object instances) without losing its focus on the local context (subject or object nodes). We define, 
\begin{align} \label{pee}
\begin{split}
\textit{pos}_{e_{ij}}{(k,k+1)} &=\left[\sin \left(\frac{p_i}{m^{\frac{2k}{d}}}\right),\cos\left(\frac{p_i}{m^{\frac{2k}{d}}}\right)\right] \\
\textit{pos}_{e_{ij}}{(k+2,k+3)} &=\left[\sin \left(\frac{p_j}{m^{\frac{2k}{d}}}\right),\cos\left(\frac{p_j}{m^{\frac{2k}{d}}}\right)\right].
\end{split}
\end{align}
Eq. \ref{pee} describes positional encoding for an edge, where $p_i$ and $p_j$ are the positions of the nodes $n_i$ and $n_j$, $m$ is maximum number of sequence of nodes,  $d=2048$, and $k$ denotes the $k^{th}$ position in the positional encoding features vector.

\subsubsection{Decoder E2N Attention}
\label{sub:cee}
%Perhaps the most important part in scene graph generation or detection depends on the expressiveness of the edge features and how well they can depict relations among object pairs.
For an edge $e_{ij}$ of the \textit{Edge Queries}, (between nodes $n_i$ and $n_j$), its bounding box location ($b_{ij} \in \mathbb{R}^4$) and initial visual features $v_{ij} \in \mathbb{R}^{4096}$ are derived from the union of the bounding boxes of both nodes as shown in Figure \ref{fig:atten4rel}. %Its  are obtained from  $v_$Faster-RCNN as described in \cite{zellers2018neural}. 
%Afterwards, the edge-specific binary-mask spatial feature ($e^{\textit{spt}}_{ij}$) is combined with visual features\footnote{we followed the same mask generation process from  \cite{zellers2018neural}}.
We concatenate the  GloVe vector embedding from both of its node labels $s_{ij}$ with  the previously obtained box and visual features for the semantic enrichment of the edge. Subsequently, a linear projection layer ($W_{\textit{edge}}$) is used to obtain the initial edge feature vector  ($f^{\textit{in}}_{ij}$) or the \textit{Edge Queries} as
\begin{align}
\label{ein}
 f^{in}_{ij} = W_{edge}([v_{ij};s_{ij};b_{ij}]).
\end{align}
% Next,  $f^{in}_{ij}$ is used for  accumulating context information across all nodes. 

We argue that a well contextualized edge is needed for complex global scene representation and this can only be achieved if the edge exploits an larger scene context. In traditional transformer decoder, a masked attention is used, limiting the edges attention only to a part of the sequence. The accumulation of global context for an edge requires a unique mechanism so it can preserve its local dependency while exploring global context.

% A well contextualized edge would be beneficial for the relation classification, if it can exploit a larger scene context.

% To serve this objective, we remove the masked attention mechanism from the original transformer decoder so that an edge can attend the whole sequence of nodes, not just a part of it.

% We have introduced several modifications in the original decoder \cite{vaswani2017attention}, e.g., positional encoding and E2N attention, such that the network learns relational (e.g. spatial, semantic) influences from other nodes:

% Then, the result is  forwarded to the position-wise feed forward network  as shown in Figure \ref{fig:atten4rel}. Our implicit modeling assumption is that multi-hop propagation of context between edges and nodes helps to encode necessary information about the scene or image.

Empirically we found that applying self-attention between edges does not help, as necessary context can be accumulated using N2N and E2N attention. See also  Table \ref{tab:ablation_dec1}. Hence, we have removed the edge-to-edge self-attention in our decoder. At first, E2N cross-attention has been applied from an edge to all the nodes. Finally, we get the contextual edge features ( $f^{\textit{final}}_{ij} \in \mathbb{R}^{2048}$)  as, 
\begin{align}\label{efin}
f^{\textit{final}}_{ij} =\textit{decoder}(f^{\textit{in}}_{ij} + {pos}_{e_{ij}}, f^{\textit{final}}_{i={1..N}}).
\end{align}
% \end{enumerate}
where \textit{decoder} is a stack of multi-head attention with our proposed E2N attention, positional encoding.
\subsection{Directed Relation Prediction Module (RPM)}
\label{sub:rpm}
Relation is a directional property, i.e., \textit{subject} and \textit{object} cannot be exchanged. After obtaining the context rich node and edge embeddings, an initial directed relational embedding ($rel^{\textit{in}}_{{i}\rightarrow {j}} \in \mathbb{R}^{2048 \times 3 +512}$) has been created as
\begin{align}
\label{rel1}
\setlength\itemsep{0pt}
\textit{rel}^{\textit{in}}_{{i}\rightarrow {j}} = [f^{\textit{final}}_i;f^{\textit{final}}_{ij};f^{\textit{final}}_j;GAP(I_{feature})]  .
\end{align}
 where $GAP(I_{feature}) \in \mathbb{R}^{512}$ is the global average pool of the image feature obtained from object detector.
As a next step,  $rel^{\textit{in}}_{{i}\rightarrow {j}}$ is passed through a sequential block of neural networks, which we call directed relation prediction module or $RPM$.  $RPM$ takes  ($rel^{\textit{in}}_{{i}\rightarrow {j}}$) as input and the result is  normalized with layer norm \cite{ba2016layer} followed by two blocks of linear layers and Leaky ReLU \cite{xu2015empirical} nonlinearity for the predicate classification as described in Eq. \ref{rel1}. Details on $RPM$ can be found  in the  supplementary material. We postulate that a larger embedding space ($RPM$), with normalized embeddings from nodes and edges, will effectively combine the necessary context, and  
\begin{align}
\label{rel2}
\setlength\itemsep{0pt}
\textit{rel}^{\textit{final}}_{{i}\rightarrow {j}} = RPM(rel^{\textit{in}}_{{i}\rightarrow {j}}).
\end{align}
Finally, we get the \textit{softmax} distribution over all $p$ predicate categories from the final relation vector $(\textit{rel}^{\textit{final}}_{{i}\rightarrow {j}} \in \mathbb{R}^{2048})$ through a linear layer ($W_{\textit{p}}$). Note that $p$ is the number of relation present in the dataset. We have also added the frequency baseline (\textit{fq}) from \cite{zellers2018neural} to model the dataset bias and obtain as
\begin{align}
\begin{split}
Pr(r_{{i}\rightarrow {j}}|I)  = \text{softmax}(W_{\textit{p}} \textit{rel}^{\textit{final}}_{{i}\rightarrow {j}})+\textit{fq}(\textit{i},\textit{j})). \label{rfin}
\end{split} 
\end{align}
The $Pr(r_{{i}\rightarrow {j}}|I)$ denotes the final relationship distribution among $n_i$ and $n_j$ for a given image $I$. 
\section{Experiments}
\label{sec:exp}
This section will describe the  dataset and explain implementation details of our network pipeline and spatial embedding implementation. \footnote{Code is available at:\url{https://github.com/rajatkoner08/rtn} }

\subsection{Datasets}
We have used three most commonly used scene graph dataset, i.e., Visual Genome \cite{krishna2017visual}, GQA\cite{hudson2019gqa} and VRD \cite{lu2016visual},  for our experimental evaluation.

\noindent\textbf{Visual Genome (VG)} is one of the most challenging datasets for scene graph detection and generation for real world images. The original dataset consists of 108,077 images with annotated object bounding boxes, class, and binary relations among the objects. The annotations are quite noisy: e.g.,  multiple bounding boxes are provided for a single object. To alleviate this problem, Lu et al. \cite{Xu_2017_CVPR} proposed a refined version of the dataset, which consists of the most frequently occurring 150 objects and 50 relationships. To have a fair comparison with most of the present state of art model \cite{zellers2018neural,newell2017pixels,zhang2019graphical,herzig2018mapping,zhang2019large} we have used this refined dataset. Also, our train (55K), validation (5K), and test (26K) split are the same as per the dataset.

\noindent\textbf{GQA} is one of the largest and diverse scene graph datasets consisting of 1704 classes and 311 relationship labels as proposed in \cite{hudson2019gqa}. It uses the same images from Visual Genome\cite{krishna2017visual} with more clean (e.g., more accurate spatial location) and normalizes class and relationship distribution. GQA is more challenging than other datasets as each image is annotated with a dense scene graph and a large number of relations. We have used K-fold data for training and report our result on ``val'' set mentioned in GQA. We omitted the classwise frequency distribution \cite{zellers2018neural} for GQA for two reasons, first GQA is more normalized than VG, and second due to a large number of classes and relationships present in GQA incurred a large memory overhead.

\noindent\textbf{VRD} contains 4000 training and 1000 test images with 100 objects and 70 predicate categories. We evaluate our model with the same COCO pretrained backbone as used in \cite{zhang2019graphical}. The evaluation metric is the same as \cite{lu2016visual} that report R@50 and R@100 metric for relationship, phrase, and predicate detection.

%%%%%%%%%%%%%%%%%%%%%%%%%%%%%%% VG %%%%%%%%%%%%%%%%%%%%%%%%%%%%%%%%%%%
\begin{table*}[ht!]
\centering
\footnotesize
\begin{tabular}{lccc|ccc|ccc|cc|cc|c}
\toprule
\multicolumn{1}{c}{ \textbf{Model} }    
& \multicolumn{9}{c|}{w/ Graph Constraint}  
& \multicolumn{4}{c|}{w/o Graph Constraint} 
& \multirow{3}{*}{ \textbf{Mean}}  \\
&\multicolumn{3}{c|}{\textbf{SGDET}}
&\multicolumn{3}{c|}{\textbf{SGCLS}} &\multicolumn{3}{c|}{\textbf{PRDCLS}}
&\multicolumn{2}{c|}{\textbf{SGCLS}} &\multicolumn{2}{c|}{\textbf{PRDCLS}}  \\
\multicolumn{1}{c}{\hfill\textbf{R@}}& 20                   & 50                   & 100
& 20                   & 50                   & 100             
& 20                   & 50                   & 100             
& 50 &100  & 50 & 100      &                    \\ 
\midrule
IMP   \cite{Xu_2017_CVPR} &14.6   &20.7   &24.5    & 31.7  & 34.6   & 35.4                     & 52.7  & 59.3  & 61.3                      &  43.4  & 47.2                           
&  75.2  &  83.6   
& 44.93\\
A. Emb. \cite{newell2017pixels} & 6.5  & 8.1 & 8.2 & 18.2  & 21.8  & 22.6                      & 47.9  & 54.1  & 55.4 
& 26.5 & 30.0    
& 68.0  &  75.2  
&34.04 \\
Freq.   \cite{zellers2018neural} & 17.7  & 23.5  & 27.6   & 27.7  & 32.4  & 34.0                      & 49.4  & 59.9  & 64.1                      & 40.5    & 43.7   
&71.3   & 81.2   
&44.07\\
MotifNet   \cite{zellers2018neural} & 21.4  & 27.2  & 30.3   & 32.9  & 35.8  & 36.5                      & 58.5  & 65.2  & 67.1                      & 44.5    & 47.7   
&81.1   & 88.3   
&49.50\\
CMAT   \cite{chen2019counterfactual} & 22.1  & 27.9  & 31.2   & 35.9  & 39.0  & 39.8                      & 60.2  & 66.4  & 68.1                      & 48.6    & 52.0   
&82.2   & 90.1   
&50.97\\
VRU  \cite{zhang2019large}    & 20.7 & 27.9 & 32.5   & 36.0  & 36.7  & 36.7                      & 66.8  & 68.4  & 68.4                      &  -  &  -   
&  -  &  -   
&52.16\\
KREN  \cite{chen2019knowledge}    & - & 27.1 & 29.8   & -  & 36.7  & 37.4                      & -  & 54.2  & 59.1                      &  45.9  &  49.0   
&  81.9  &  88.9   
&51.00\\
ReIDN   \cite{zhang2019graphical}  &21.1  &28.3   &32.7      & 36.1  & 36.8  & 36.8                      & 66.9  & 68.4  & 68.4                      &  48.9  & 50.8                          
&  93.8  &  97.8   
&52.83\\ 
GPS Net   \cite{lin2020gps}  &22.3  &28.9   &\textbf{33.2}      & 41.8  & 42.3  & 42.3                      & 67.6  & \textbf{69.7}  & \textbf{69.7}                      &  -  & -                          
&  -  &  -   
&46.42\\ 

\midrule
% \textbf{RTN} with Norm Loss    &-  &-   &-      & 41.9  & 42.8  & 42.8                      & 68.1  & 68.4  & 68.4                      &  -  & -                          
% &  -  &  -   
% &55.4\\ 
{\bfseries}\textbf{RTN (Ours)}  
& \textbf{22.5}  & \textbf{29.0} & 33.1 & \textbf{43.8}  & \textbf{44.0} & \textbf{44.0}  & \textbf{68.3}        
& 68.7        & 68.7  
& \textbf{61.3}   & \textbf{62.3}
& \textbf{97.2}    & \textbf{99.1}          & \textbf{57.1}\\ 
\bottomrule\\
\end{tabular}
\caption{Comparison of our model with state-of-the-art methods tested on  Visual Genome \cite{krishna2017visual}. We have computed mean performance, 
as some of the works did not provide all scores.  Evaluation with \lq Graph constraint\rq allows only one relation between an object pair,
and  ``No Graph constraint'' allows multiple relations between an object pair.}
\label{tab:results}
\end{table*}

%%%%%%%%%%%%%%%%%%%%%%%%%%%%%%% GQA %%%%%%%%%%%%%%%%%%%%%%%%%%%%%%%%%%%
\begin{table}
\centering
\setlength{\tabcolsep}{2pt}
\begin{tabular}{llll|lll|l} 
\toprule
Model               & \multicolumn{3}{c|}{\textbf{SGCLS}}                            & \multicolumn{3}{c|}{\textbf{PREDCLS}}                   & \textbf{Mean}  \\
\multicolumn{1}{c}{\hfill\textbf{R@}} &20 &50  &100 &20 &50 &100. &\\ 
\midrule
IMP\cite{Xu_2017_CVPR}                 &6.3            &9.4                       &11.2            & 47.3           & 69.8          & 81.4          & 37.6      \\
Neural Motif\cite{zellers2018neural}        & 6.5         & 9.9                  & 11.9         & 51.2           & 73.6          & 84.2          & 39.6      \\
Unbiased TDE\cite{tang2020unbiased} &5.8            &8.8                       &10.6                & 51.6           & 74.0          & 84.6          & 39.2      \\ 
\midrule
\textbf{RTN(ours)}                 & \textbf{9.2} & \textbf{11.9}          & \textbf{12.2} & \textbf{55.3} & \textbf{78.2} & \textbf{88.1} & \textbf{42.3}      \\
\bottomrule
\end{tabular}

\caption{Comparison of our model on the GQA dataset for predicate and scene graph classification. Mask-RCNN\cite{he2017mask} is used as backbone object detector fine tuned in GQA. }
\label{tab:gqa}
\end{table}

%%%%%%%%%%%%%%%%%%%%%%%%%%%%%%% Mean Recall %%%%%%%%%%%%%%%%%%%%%%%%%%%%%%%%%%%
\begin{table}[t!]
\centering
\small
\begin{tabular}{lc|c}
\toprule
\multirow{2}{*}{\textbf{Model}} & \textbf{SGCLS}       & \multicolumn{1}{l}{\textbf{PREDCLS}}  \\

\multicolumn{1}{c}{\hfill\textbf{mR@}} & \multicolumn{1}{c|}{100} &\multicolumn{1}{c}{100}\\
% \textbf{\phantom{abcd}mean Recall@}   & 100  &100\\
%  & mean-Recall@100  & mean-Recall@100\\
\midrule
VG &  &\\
\cmidrule(lr){1-1} 
IMP \cite{Xu_2017_CVPR} & 6.0 &10.5\\
FREQ \cite{zellers2018neural} & 8.5 &16.0\\
MotifNet\cite{zellers2018neural}& 8.2 &15.3\\
KERN\cite{chen2019knowledge}& 10.0 &19.2\\
VCTREE-HL\cite{tang2018learning}& 10.8 &19.4\\
GPS-Net\cite{tang2018learning}& \textbf{12.6} &\textbf{22.8}\\
%\midrule
% {\bfseries}\textbf{Relation Transformer} with Norm Loss
% & \textbf{13.1} 
% & 21.8\\
{\bfseries}\textbf{RTN (Ours)}
& \textbf{12.6} 
& 20.3\\
\midrule
GQA &  &\\
\cmidrule(lr){1-1}
IMP \cite{Xu_2017_CVPR} & 0.5 &2.2\\
MotifNet\cite{zellers2018neural}& 0.8 &2.9\\
Unbiased TDE\cite{tang2020unbiased}& 0.7 &2.8\\
{\bfseries}\textbf{RTN (Ours)}
& \textbf{1.4} 
& \textbf{4.5}\\
\bottomrule\\
\end{tabular}
\caption{Comparison on the mean-Recall@100 metric between various methods across all the 50  and 311 relationship categories for VG and GQA correspondingly.}
\label{tab:meanRecall}
\end{table}

%%%%%%%%%%%%%%%%%%%%%%%%%%%%%%% VRD %%%%%%%%%%%%%%%%%%%%%%%%%%%%%%%%%%%
\begin{table}[t]
\centering
\small
\begin{tabular}{lcc|cc}
\toprule
\multirow{2}{*}{ \textbf{\phantom{abcdef}Model} }
&\multicolumn{2}{c|}{\textbf{Relation}} &\multicolumn{2}{c}{\textbf{Phrase}}\\
&\multicolumn{2}{c|}{\textbf{Detection}} &\multicolumn{2}{c}{\textbf{Detection}}\\
\multicolumn{1}{c}{\hfill\textbf{R@}} &50 & 100 &50 &100\\ 
\midrule
VTransE \cite{zhang2017visual} & 19.4 &22.4& 14.1 &15.2\\
Vip-CNN \cite{li2017vip} & 17.3 &20.0 & 22.8 &27.9\\
KL distilation\cite{yu2017visual}& 19.2 &21.3 & 23.1 &24.0\\
Zoom-Net\cite{yin2018zoom}& 18.9 &21.4 & 24.8 &24.1\\
RelDN\mbox{*} \cite{zhang2019graphical}& 25.3 &28.6 & 31.3 &36.4\\
GPS-Net\mbox{*} \cite{lin2020gps}& 27.8 &31.7 & \textbf{33.8} & \textbf{39.2}\\
\midrule
{\bfseries}\textbf{RTN (Ours)}\mbox{*}
& \textbf{28.1}
& \textbf{32.0} & 33.5 &38.7\\
\bottomrule\\
\end{tabular}
\caption{Comparison of our model on the VRD dataset for relation detection and Phrase Detection. \mbox{*} uses the same detector pretrained on COCO.}
\label{tab:vrd}
\end{table}

\subsection{Implementation Details}
We have implemented our model in PyTorch and trained it in a single Nvidia RTX 3900 GPU. We have trained the network for 20 epochs; it took approximately two days to train. 
The input to our model is an image with a size of $592 \times 592$ pixels, as in  \cite{zellers2018neural}. 
As mentioned before, 
the encoder and the decoder modules accept  input features of size 2048. We have used 3 encoder layers, 2 decoder layers and 12 attention heads for our network. Our model is optimized by SGD with momentum. A learning rate of $10^{-3}$ and batch size of 16 has been used. We have used cross-entropy loss for both of our object and relation classification loss.\footnote{List of all hyper-parameters are given in the supplementary material.} 
In training, we used one  foreground  edge (contain at least one ground truth relation) for 4 background edges (without any relation), and randomly flip some images as part of data augmentation. We have followed the same evaluation as in current benchmarks \cite{zhang2019graphical} and computed  scene graph classification (SGCLS) and predicate classification (PREDCLS). 

For scene graph detection (SGDET), we have taken the top 64 object label proposals from the object detector for each image after performing the non-maximal suppression (NMS) with intersection over union (IoU) of 0.3, as similar to \cite{zellers2018neural}. To reduce computational load in relation classification, we have only considered those pairs of nodes whose bounding boxes are overlapping.

To have a fair comparison on evaluating on visual genome, we have used Faster-RCNN \cite{ren2015faster} with VGG16 \cite{simonyan2014very} backbone pretrained on visual genome dataset as per \cite{zellers2018neural,zhang2019graphical}. We have trained Mask-RCNN\cite{he2017mask} on GQA for the backbone object detector as per \cite{tang2020unbiased}.

\section{Results and Discussion}
\label{sec:results}

\subsection{Quantitative Results}
\label{sub:quant}

For VG dataset, Table \ref{tab:results} shows the performance of our method in comparison with other methods. It clearly demonstrates that our novel context propagation for both objects and edges significantly improves most of the performance metrics. Note that in training, we only used simple cross-entropy loss in contrast to the recent literature, such as, contrastive or triplet loss (ReIDN \cite{zhang2019graphical}, VRU \cite{zhang2019large}) and Node-Priority-Sensitive (NPS) loss (\cite{lin2020gps}).
As shown under \lq No graph constraint\rq\ accuracy of $99.1\%$ R@100 in PREDCLS indicates, our model is superior than the competing models in learning the most likely relations. One of the key observations we made from Table \ref{tab:results} is  that our approach significantly improves on the R@20 metric, compared to previous state of the art RelDN \cite{zhang2019graphical} with maximum improvement of 7.7\% on SGCLS. Our model achieves slightly lower score than our contemporary \cite{lin2020gps} work on R@50 and R@100 for PREDCLS. We suspect that the frequency softening method by \cite{lin2020gps} better captures the data imbalance and long tailed distribution of relationship, than ours in those cases. However, in this study we are not focusing on loss function instead of looking deep into context propagation. And hence, we leave the class imbalance study for future work.
\begin{figure*}[ht]
\centering
    \begin{subfigure}[b]{0.24\linewidth}
        \centering
        \includegraphics[width=\linewidth, height=4cm]{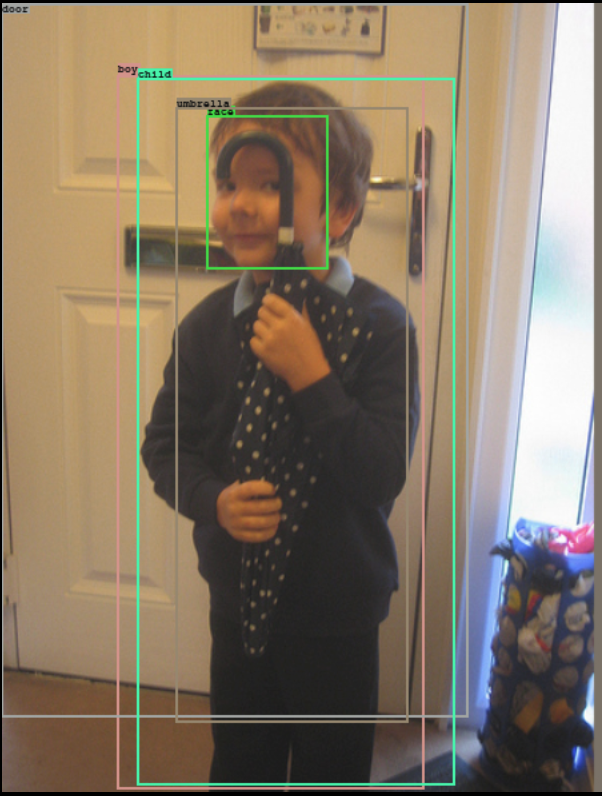}
        \caption{Image}
    \end{subfigure}    
    \begin{subfigure}[b]{0.24\linewidth}
        \centering
        \includegraphics[width=1\linewidth]{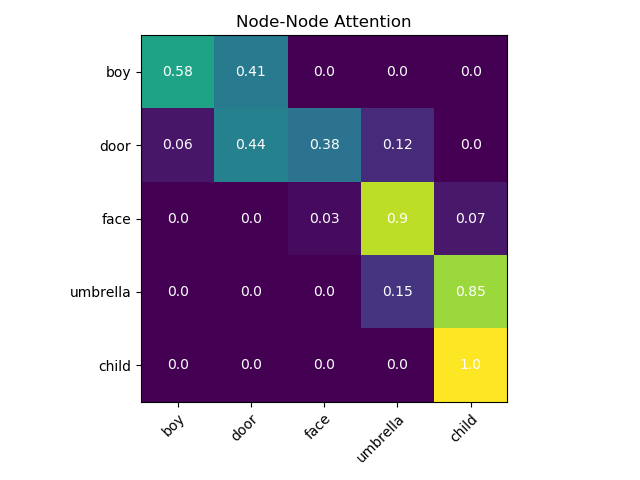}
        \caption{N2N Attention heatmap}
    \end{subfigure}
    \begin{subfigure}[b]{0.24\linewidth}
        \centering
        \includegraphics[width=\linewidth]{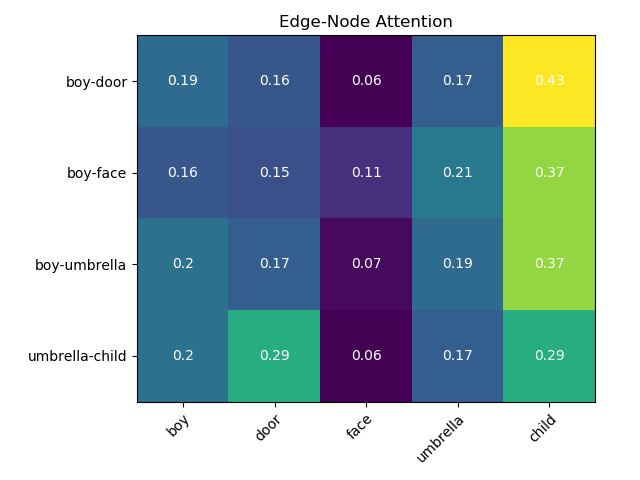}
        \caption{E2N Attention heatmap}
    \end{subfigure}
    \begin{subfigure}[b]{0.24\linewidth}
        \centering
        \includegraphics[width=\linewidth]{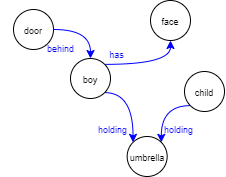}
        \caption{Generated scene graphs}
    \end{subfigure}

\caption{A positive example outputs from our network with associated attention map and scene graph.}
\label{fig:q_pos_1}
\end{figure*}

% One of the possible reasons could be that the  cross entropy loss is subjected to frequency bias \cite{Knyazev2020} and the huge dataset bias present in the Visual Genome dataset, where as \cite{lin2020gps} employs NPS loss and frequency softening for class imbalance. Recently, few other works tries to address this problem in SGG, \cite{Knyazev2020} proposed a normalized loss to deal with class imbalance present in the dataset both for foreground and background relations. Due to scope of the paper, we left the class imbalance study for future work.

% Inspired from these observations
% and to demonstrate the ability of our model on dealing with class imbalance, we have conducted an initial experiment with normalized loss (Relation Transformer with Norm loss) using the  default parameters from \cite{Knyazev2020}. The improved results of our model with normalized loss in Table \ref{tab:meanRecall} shows that  our architecture could potentially be used in combination with other special loss functions,  in the case of  highly imbalanced datasets. As a disadvantage, 
% this normalized loss hurts the performance on evaluation setting of Table \ref{tab:results}. A probable reason could be that the default parameters as specified in \cite{Knyazev2020} are unsuitable here,  and we need to  optimize hyper-parameters. We will explore this issue in future work. 

% In future we will investigate more on it.  

Table \ref{tab:gqa}, shows the performance of RTN on GQA dataset. Here RTN outperforms all competitive models by mean \textbf{3.1\%}. Compare to VG, GQA contains six times more relationship classes and three times more dense scene graph annotation. Thus GQA is more challenging for efficient propagation of context that depicts the full scene. As GQA has more normalized relationship distribution, class wise frequency bias is nonessential. Our proposed transformer based multi-hop context propagation is particularly well suited for extracting important context of a given edge query out of large number of nodes and edges. Hence, performance on GQA shows our RTN is superior over the state-of-the-art methods. Especially performance on ``PREDCLS'' depicts the effectiveness of our model on utilizing proper context information in the presence of large number of nodes or edges or in a dense scene graph.

In order to gain deeper insight on class imbalance, we computed mean-recall that takes the mean across all the relationship classes. Table \ref{tab:meanRecall} shows the performance of our model with other competing methods for the mean recall. Our RTN performs better than most of them in ``SGCLS'' and ``PREDCLS''. However, the performance on ``PREDCLS'' R@100 in VG dataset is marginally lower compared to \cite{lin2020gps} because of the class imbalance. Since GQA is more normalized than VG, thus it has less prominent class imbalance problem. We have achieved state of the art result on mean recall across all metrics on GQA.

Table \ref{tab:vrd}, shows the performance of our network on the VRD dataset; for a fair comparison, we have used the same object detector as in \cite{zhang2019graphical}. Here also, we have performed better than or on par with other state-of-the-art models on relations classification. Transformer based architectures are reported to require a large number of training data, but despite of relatively small sample size of VRD, our node and edge based context propagation helps RTN for an efficient accumulation of scene context.
% and almost similar to our contemporary work \cite{lin2020gps} on phrase classification. 

\subsection{Qualitative Results}

Figure. \ref{fig:q_pos_1}, shows a positive example with N2N and E2N attention. Here N2N attention shows how the presence of an object influence other like boy-door, child-umbrella etc. E2N attention heat map shows the importance of surrounding nodes for a particular edge (e.g child, boy play an important role in most of relation as they are same semantic object ).  More  positive and negative examples can be found  in the supplementary section. 

Additionally, an analysis of the  errors of the Relation Transformer network provides insight into what the network has learned. For example, \lq on\rq\ is the most mispredicted relation in our evaluation settings for Visual Genome. The relationship \lq on\rq\ is falsely predicted as \lq of\rq\ for 56.9\% out of total false predictions among all relationships. Interestingly, for most of the examples like \lq \textit{Face, of, Woman} \rq\ is more appropriate than \lq \textit{Face, on, Woman} \rq\ , indicating that  the network is not necessarily failing to predict correctly,  rather it is predicting more suitable or  semantically similar relationships. A probable reason for these false predictions is the significant bias in the training dataset. 
\subsection{Ablation Study}
% To show the efficacy of our proposed novel object (N2N), edge context (E2N) enrichment module and directed RPM,
In the following, we present the results of three ablation experiments in order to demonstrate the efficacy of our proposed modules and the selection of hyper-parameters on VG dataset.

\subsubsection{Encoder-Decoder Layers and RPM}
\label{sub:ablation1}
First, we compare the performance of the N2N and E2N modules for varying number of layers in combination with the RPM. The results are presented in the following Table \ref{tab:context}. We make the following observations.

\begin{table}[!ht]
\centering
\small
\begin{tabular}{l|l|l|c@{\hskip 0.2in}c@{\hskip 0.2in}c}
\toprule
\multirow{2}{*}{\textbf{\#N2N}} & \multirow{2}{*}{\textbf{\#E2N}} &
\multirow{2}{*}{\textbf{RPM}} &
\multicolumn{3}{l}{\textbf{\phantom{abcdefg}PREDCLS}} \\
                                         &   &                                     & \textbf{R@20}    & \textbf{R@50}   & \textbf{R@100}   \\
\midrule
1 &1 & yes &66.7  &67.1 &67.1 \\ 
1 &2 & yes &68.0  &68.4 &68.4 \\
2 &1 & yes &67.4  &67.8 &67.8 \\ 
2 &2 & yes &68.1  &68.5 &68.5 \\ 
\textbf{3(ours)} &\textbf{2(ours)} & yes  &\textbf{68.3}  &\textbf{68.7} &\textbf{68.7} \\
3 &2 & no &67.8  &68.2 &68.3\\
3 &3 & yes &68.2  &68.5 &68.5\\
\bottomrule
\end{tabular}\\
\caption{Ablation study of our proposed N2N, E2N and relation prediction module on Visual Genome for PREDCLS.}
\label{tab:context}
\end{table}

\begin{enumerate}
    \item All nodes need to be contextualized well enough, in order to propagate context for edges. This is demonstrated in rows 4 and 5 of Table \ref{tab:context}. Adding more edge context layers without adequate object context propagation limits the performance. Thus, it shows the importance of contextualized objects.
    \item After objects are properly contextualized, adding the optimal number of edge-context modules on top of object-context modules increases the performance. Jointly it shows the  importance of both  modules (rows 4 and 5).  
    \item If we use a large number ($\geqslant 3$) of E2N and N2N layers, then the performance decreases. One of the possible reason is that, as the network size grows, it becomes hard to optimize the network (row 7).
    \item Directed RPM facilitates the forming of the necessary embedding space after completion of optimal contextualization of nodes and edges. If we replace the RPM module with a simple linear layer without any normalization, it hurts the performance (row 6).
    %\item We have also conducted experiments without using any E2E attention in our final configuration (3 object context, 2 edge context layer), and the result did not change significantly. We assume N2N and E2N provides sufficient contextualization. This remains an open question and requires further analysis.
\end{enumerate}
From these observations, we can clearly infer that all three modules have significantly contributed to relationship classification.

%%%%%%%%%%%%%%%%%%%%%%%%%%%%%%%%%%%
\subsubsection{Decoder E2N Attention and Positional Encoding}
Next, we present necessary ablation experiments in support of the benefits of the proposed changes in the decoder architecture as discussed in Section \ref{sec:intro}. We have conducted three experiments to understand the impact of these changes as presented in Table \ref{tab:ablation_dec1}. We make the following observations.

\begin{table}[!ht]
\centering
\small
\begin{tabular}{lccc}
\toprule 
      \multicolumn{1}{c}{\textbf{Model}} &
      \multicolumn{2}{c}{\textbf{\phantom{abcd}PRDCLS}} \\
    \multicolumn{1}{c}{\hfill\textbf{R@}}&20         & 50         & 100       \\
\midrule
with \cite{vaswani2017attention}'s decoder           & 66.5       & 67.1       & 67.1      \\
with only E2N attention                              & 67.6       & 68.0       & 68.0      \\
\midrule
{\bfseries}with only E2N attention  & & &\\
+ proposed PE \textbf{(RTN)} & \textbf{68.3}       & \textbf{68.7}       & \textbf{68.7}   \\
\bottomrule\\
\end{tabular}
\caption{Ablation study of our proposed changes in original transformer decoder \cite{vaswani2017attention}, like changes in positional encoding, removal of decoder-decoder attention on Visual Genome dataset.}
\label{tab:ablation_dec1}
\end{table}

\begin{enumerate}
    \item \textit{With \cite{vaswani2017attention}'s decoder:}  shows the network performance without any modification from \cite{vaswani2017attention}'s transformer decoder. In this setting, at first,  the decoder applies attention across all edges (decoder self-attention), afterward,  from edge to all nodes (decoder-encoder cross attention or ours E2N attention). However, for SGG task, primary context of an edge should come from the immediate and neighboring nodes (E2N attention). Hence, self-attention before cross-attention, as in \cite{vaswani2017attention}'s decoder, hinders node to edge context propagation and limits performance.
    % Theoretically,  for SGG, primary importance of an edge is (1) to accumulate context from the nodes (E2N), (2) from other edges, not the reverse way.
    
    % \item  \textit{with reordered decoder: } shows that network performances increased with re-ordering (E2N$\rightarrow$decoder-decoder) of attention. Here we didn't change the positional encoding for the original decoder. The result clearly demonstrates our claim that,  at first,  an edge needs to explore all node context, before it focuses on the context from  other edges.
    
    \item \textit{With only E2N attention:} shows our network performance with only E2N attention (without self-attention in the decoder) and position-wise feed forward network. Here we didn't change the positional encoding for the decoder. Removing the self-attention in decoder, improves the performance. One possible reason for this could be easy accumulation of necessary context from the nodes needed for inferring relation of the image.
    
    \item \textit{With only E2N attention + proposed PE (RTN):} shows our Relation Transformer model. Here we have applied our proposed positional encoding, and the results show that the preservation of 
    the source node identity of an edge helps the network efficiently accumulate the global context without losing the local context.
\end{enumerate}
Our experiment shows that our design changes in the original transformer architecture are effective and helpful for relation classification.

%%%%%%%%%%%%%%%%%%%%%%%%%%%%%%%%%%%
\subsubsection{Transformer and RPM Feature Space}
Finally, we report results on ablation studies to analyze the impact of feature embeddings in transformer inputs i.e., semantics or GloVE and spatial embedding. Additionally, we perform ablation on the RPM feature space i.e., global average pooling and frequency bias. Both the ablation studies are presented in Table \ref{tab:ablation_dec2}. We make the following observations.
\begin{table}[!ht]
\centering
\small
\begin{tabular}{lccc}
\toprule 
      \multicolumn{1}{c}{\textbf{Model}} &
      \multicolumn{2}{c}{\textbf{\phantom{abcd}PRDCLS}} \\
\textbf{\phantom{abcd}R@}     &20         & 50         & 100       \\
\midrule
without word embedding      & 67.7       & 68.1       & 68.1      \\
without spatial embedding   & 68.2       & 68.6       & 68.6      \\
without Global Avg Pool     & 68.0       & 68.4       & 68.4      \\
without Freq Bias           & 66.7       & 67.0       & 67.0      \\
\midrule
{\bfseries} with every features & \textbf{68.3}       & \textbf{68.7}       & \textbf{68.7}   \\
\bottomrule\\
\end{tabular}
\caption{Ablation study of our proposed changes in original transformer decoder \cite{vaswani2017attention}, like changes in positional encoding, removal of decoder-decoder attention on Visual Genome dataset.}
\label{tab:ablation_dec2}
\end{table}

\begin{enumerate}
    \item Without word embedding, we see a decrease in recall. We attribute this to the fact that word embedding provides additional semantic cues based on the correlated object classes, which is helpful for the SGG task.
    \item Without spatial embedding of the bounding box coordinates, the performance sufferers, but the drop is marginal. For an accurately known global context of an object, spatial embedding provides primarily redundant information. However, when the object detector does not accurately localize the object, spatial embedding can provide additional information to compensate for it.
    \item Without Global Avg Pool, we also observe a performance drop. We suspect a lack of sufficient global image information behind the decrease in recall. With Global Avg Pool, it is easier for the network to aggregate the context of individual nodes and edges that are far apart in spatial location.
    \item Without frequency bias, we see significant performance drops. This is because of the high-class imbalance in the VG dataset, which dominates the loss function. Thus frequency bias is a crucial part of the RPM to mitigate the issue of class imbalance.
\end{enumerate}

\section{Conclusion}
In this paper, we propose a transformer view on the scene graph generation task. Notably, we explore node-to-node and edge-to-node association through the lens of self-attention and cross-attention, respectively. The resultant node and edge embedding from the transformer encoder and decoder, respectively, contributes to robust and discriminate relation classification through our relation prediction module. We achieve a consistent improvement over the state-of-the-art models in the evaluation metrics across small, medium, and large-scale relational datasets. Our modular nature of RTN can offer contextualized node and edge features for other various tasks, such as visual question answering, text to image generation, etc. Although our work's current focus remains on attention-based supervised relational context modeling, future work can be carried upon solving class imbalance issues and end-to-end self-supervised relational feature learning.

% In conclusion, we see that each component of our embedding contributes to the rich discriminative feature selection, which justifies our proposed embedding space.

% In our qualitative results, we provide a detailed analysis of N2N and E2N attention heatmap and how a node influences other nodes and edges for relation classification. 

% Finally, our approach employs  direction-aware joint relationship classification of nodes and edges. This modular network can easily be extended for other downstream tasks based on  the contextualized node and edge embeddings.

%\begin{acknowledgements}
%\end{acknowledgements}


\bibliographystyle{spmpsci}      % mathematics and physical sciences
\bibliography{bibliography}   % name your BibTeX data base
\clearpage
\appendix
% \documentclass[a4paper,10pt]{article}
% \usepackage[utf8]{inputenc}

% \usepackage{graphicx}
% \usepackage{comment}
% \usepackage{amsmath,amssymb} % define this before the line numbering.
% \usepackage{booktabs, floatrow, makecell}
% \usepackage{multirow}
% \usepackage{color}
% \usepackage{subcaption}
% \usepackage[pagebackref=true,breaklinks=true,letterpaper=true,colorlinks,bookmarks=false]{hyperref}

% \title{Relation Transformer Network\\
% \large Supplementary Material }

% \begin{document}
% \maketitle
\normalsize
% This is a supplementary material for our paper \lq Relation Transformer Network\rq. Here, we will discuss more about implementation details, attention map and qualitative results. 

% \subsubsection{Spatial Embedding}\label{se}
% We postulate that relationships among two objects also depend on their spatial location. For example in  Figure \ref{fig:atten4rel}, the relation (\textit{sitting on}) among apple (in red box) and plate (in black box) can be inferred through spatial location. We have encoded spatial information using the normalize position \cite{ren2015faster} of nodes and edges, 
% and spatial masks \cite{zellers2018neural,dai2017detecting} of subject and object nodes. A normalized coordinate features of $b_i$ and union of two bounding boxes ($b_{ij}$) can be expressed as,
% \begin{align}
% b_{\text{norm}}=(\dfrac{x}{w_{img}},\dfrac{y}{h_{img}},\dfrac{x+w}{w_{img}},\dfrac{y+h}{h_{img}},\dfrac{wh}{w_{img}h_{img}}) 
% \end{align}
% where bounding boxes are provided in the format $(x, y, w, h)$, and $w_{img}$,$h_{img}$ are the width and height of the image, and $(x,y)$ are the left most corner of the bounding box and ($w,h$) are their corresponding width and height. To leverage more on spatial embedding, we have used a binary mask of two boxes $b_i$ and $b_j$ and fed them  to a convolutional layer specified in \cite{zellers2018neural}. Afterwards this spatial features was added with edge visual features ($e^{\textit{vis}}_{ij}$).

\section{Implementation Details}
In this section we will list out hyper-parameter used in final model.
\begin{enumerate}
    \item optimizer : Stochastic Gradient Descent(SGD)
    \item learning rate : $10^{-3}$ with reduce on plateau and patience $3$
    \item batch size : 16
    \item dropout \cite{srivastava2014dropout} : $0.25$
    \item Context Propagation of Objects : 3 E2N modules
    \item Context Propagation for Edges : 2 E2N modules
    \item weight initialization : Xavier normal
    \item attention head : 12 attention heads are used in both N2N and E2N.
    \item random seed : 42
    \item Directed Relation Prediction Module (RPM): As discussed in paper, a RPM module leverages upon context rich nodes ($f_i^{final},f_j^{final}$) and undirected edges ($f_{ij}^{final}$) to produce final directed relation embedding between two nodes ($rel_{i \rightarrow j}^{final}$). The input to RPM (($rel_{i \rightarrow j}^{in}$)) is normalized by LayerNorm \cite{ba2016layer}then followed by a linear layer ($W_{1} \in \mathbb{R}^{4096} $), dropout then another linear layer ($W_{2} \in \mathbb{R}^{2048} $) and finally followed by Leaky ReLU non-linearity.
    
\end{enumerate}

\section{Qualitative Results}
This section will provide a few more qualitative samples generated by our network in both positive and negative scenarios. To improve visibility and interpretability, we only consider the interaction among ground truth objects and relations in these examples.

\begin{figure*}[ht]
\centering
    \begin{subfigure}[b]{0.24\linewidth}
        \centering
        \includegraphics[width=\linewidth, height=4cm]{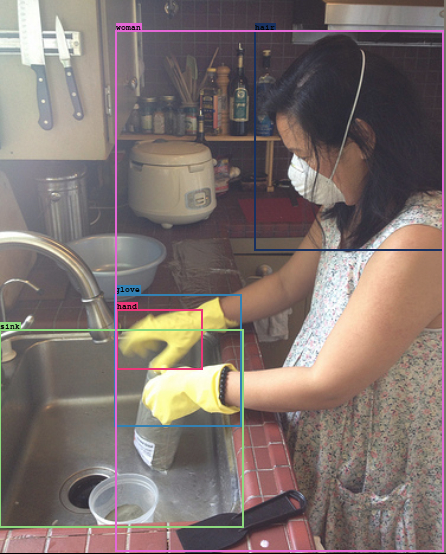}
    \end{subfigure}    
    \begin{subfigure}[b]{0.24\linewidth}
        \centering
        \includegraphics[width=1\linewidth]{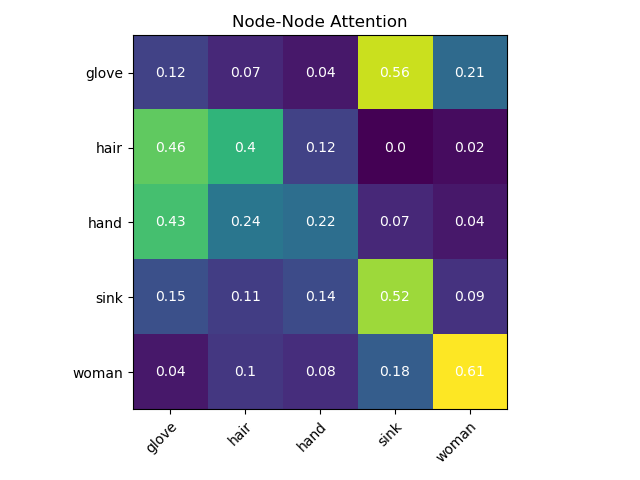}
    \end{subfigure}
    \begin{subfigure}[b]{0.24\linewidth}
        \centering
        \includegraphics[width=\linewidth]{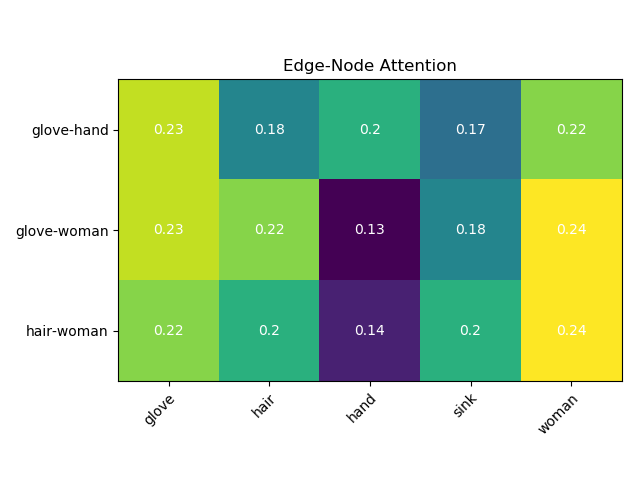}
    \end{subfigure}
    \begin{subfigure}[b]{0.24\linewidth}
        \centering        \includegraphics[width=0.7\linewidth]{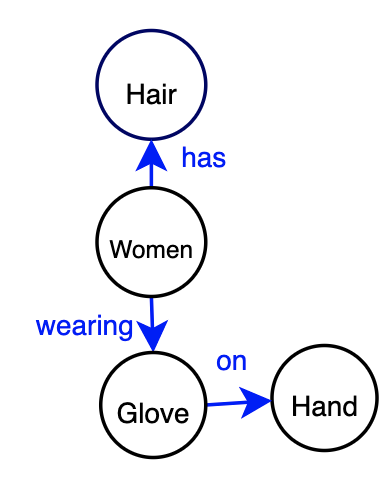}
    \end{subfigure}
    \begin{subfigure}[b]{0.24\linewidth}
        \centering
        \includegraphics[width=0.7\linewidth, height=4cm]{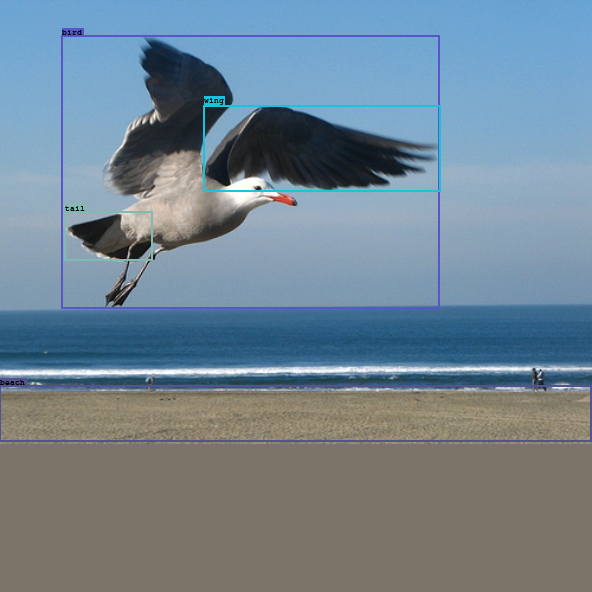}
    \end{subfigure}    
    \begin{subfigure}[b]{0.24\linewidth}
        \centering
        \includegraphics[width=1\linewidth]{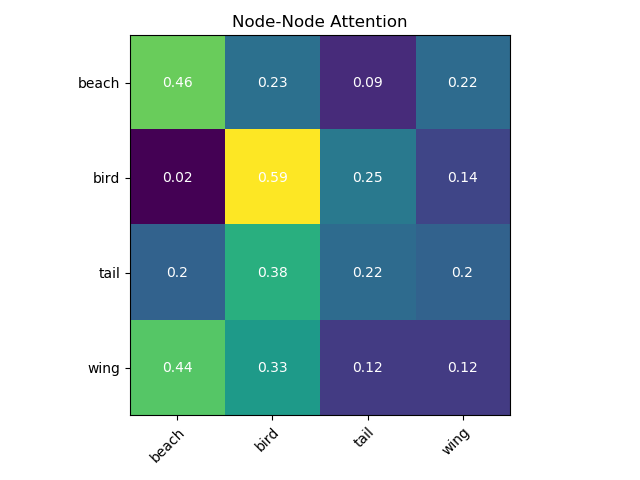}
    \end{subfigure}
    \begin{subfigure}[b]{0.24\linewidth}
        \centering
        \includegraphics[width=\linewidth]{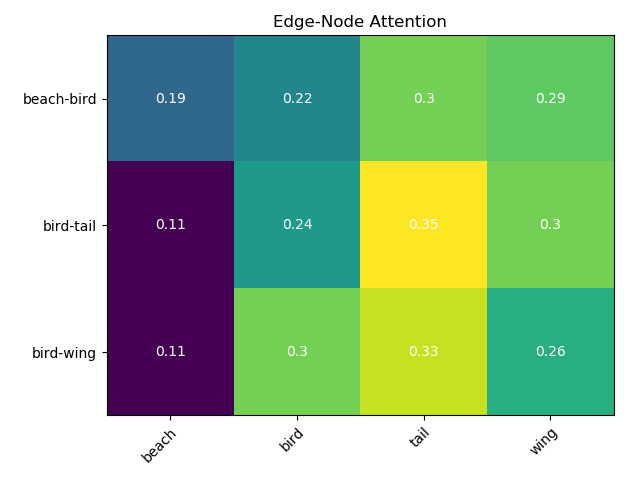}
    \end{subfigure}
    \begin{subfigure}[b]{0.24\linewidth}
        \centering
        \includegraphics[width=0.8\linewidth]{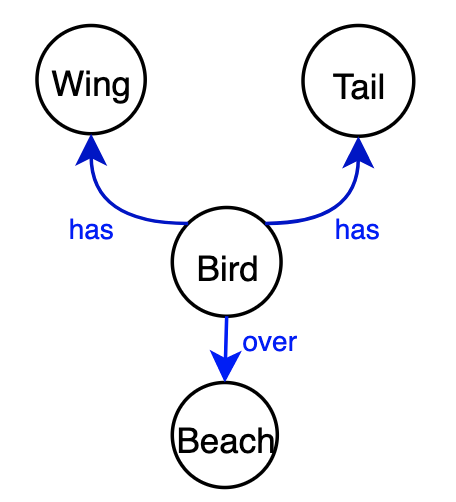}
    \end{subfigure}
    \begin{subfigure}[b]{0.24\linewidth}
        \centering
        \includegraphics[width=\linewidth, height=4cm]{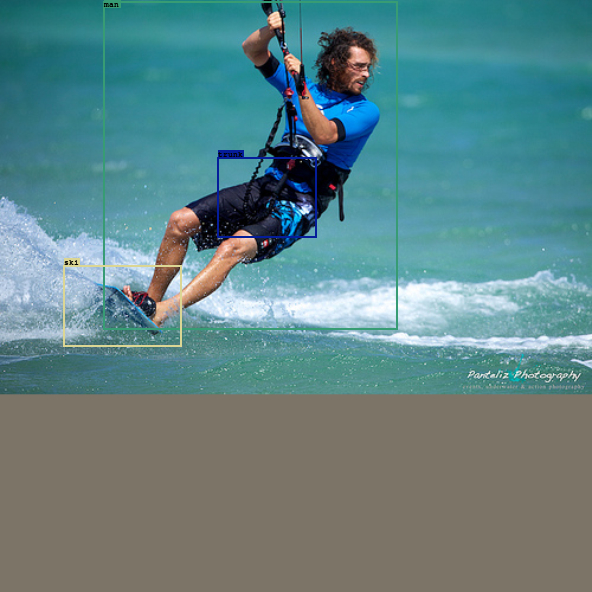}
    \end{subfigure}    
    \begin{subfigure}[b]{0.24\linewidth}
        \centering
        \includegraphics[width=1\linewidth]{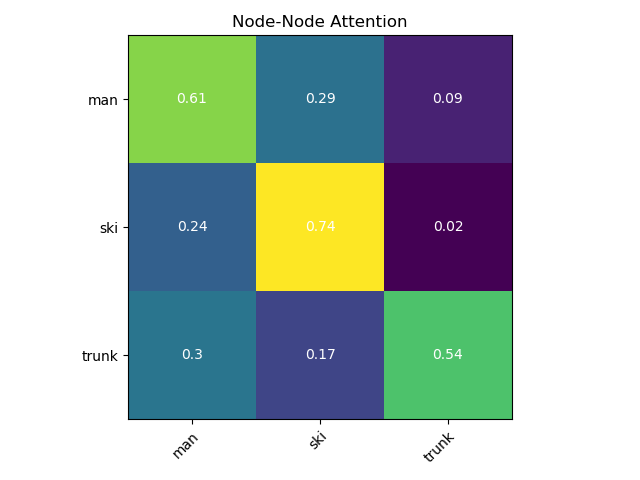}
    \end{subfigure}
    \begin{subfigure}[b]{0.24\linewidth}
        \centering
        \includegraphics[width=\linewidth]{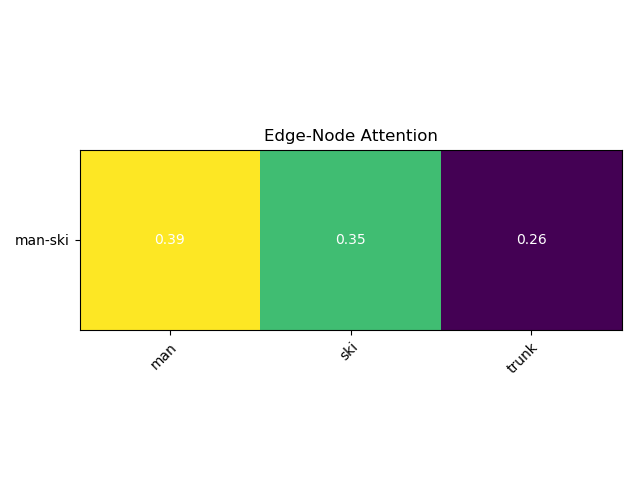}
    \end{subfigure}
    \begin{subfigure}[b]{0.24\linewidth}
        \centering
        \includegraphics[width=0.4\linewidth]{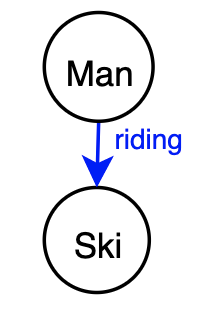}
    \end{subfigure}
    \begin{subfigure}[b]{0.24\linewidth}
        \centering
        \includegraphics[width=\linewidth, height=4cm]{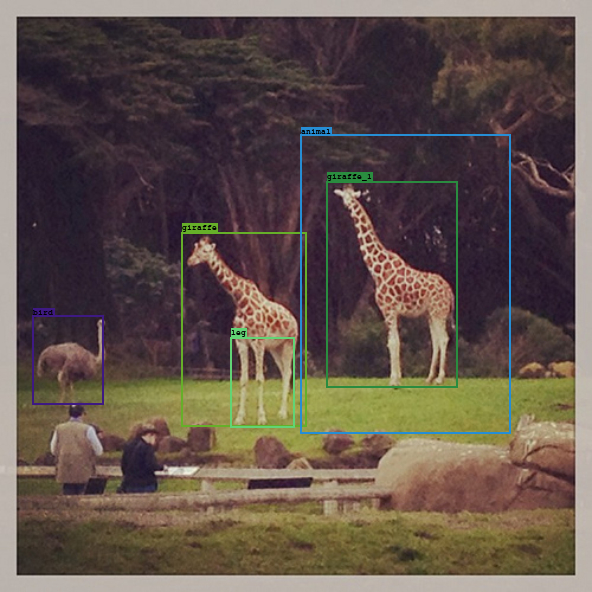}
        \caption{Image}
    \end{subfigure}    
    \begin{subfigure}[b]{0.24\linewidth}
        \centering
        \includegraphics[width=1\linewidth]{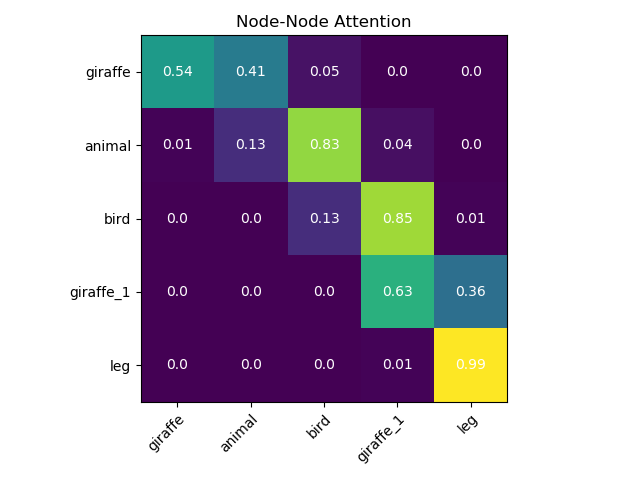}
        \caption{N2N Attention heatmap}
    \end{subfigure}
    \begin{subfigure}[b]{0.24\linewidth}
        \centering
        \includegraphics[width=\linewidth]{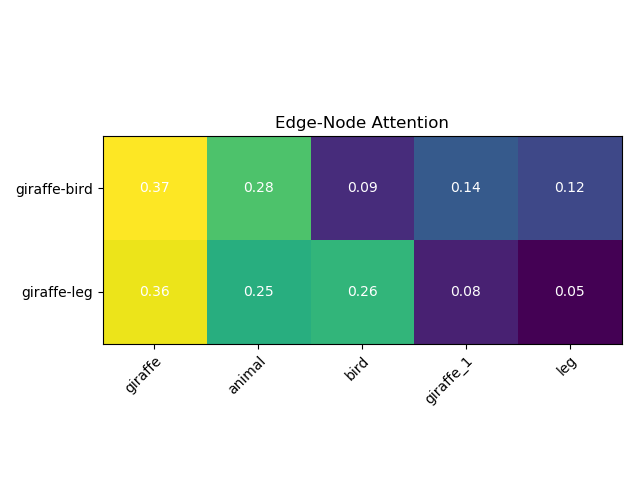}
        \caption{E2N Attention heatmap}
    \end{subfigure}
    \begin{subfigure}[b]{0.24\linewidth}
        \centering
        \includegraphics[width=0.4\linewidth]{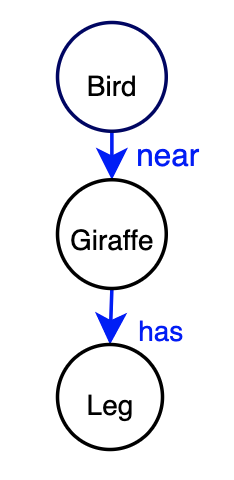}
        \caption{Generated scene graphs}
    \end{subfigure}

\caption{Some positive example outputs from our network with associated attention map and scene graph.}
\label{fig:attention map positive}
\end{figure*}

Fig. \ref{fig:attention map positive}, shows the positive scenario, where our network is able to detect correct relationships label despite the presence of repetitive bounding box (boy and child) or similar objects (giraffe). Thus, it shows the robustness of the method.

\begin{figure*}[ht]
\centering
    \begin{subfigure}[b]{0.24\linewidth}
        \centering
        \includegraphics[width=\linewidth, height=4cm]{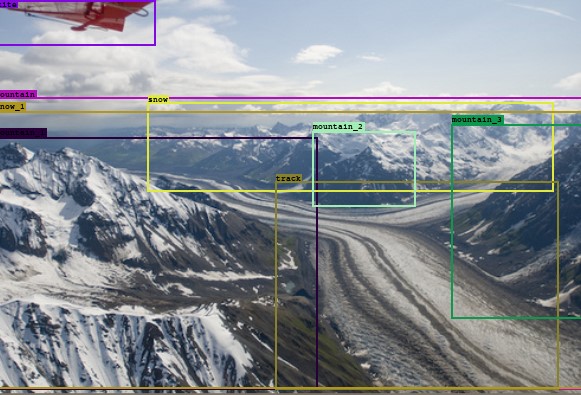}
    \end{subfigure}    
    \begin{subfigure}[b]{0.24\linewidth}
        \centering
        \includegraphics[width=1\linewidth]{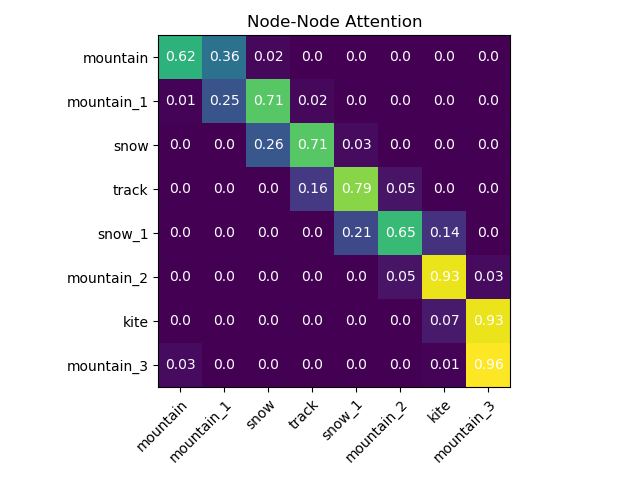}
    \end{subfigure}
    \begin{subfigure}[b]{0.24\linewidth}
        \centering
        \includegraphics[width=\linewidth]{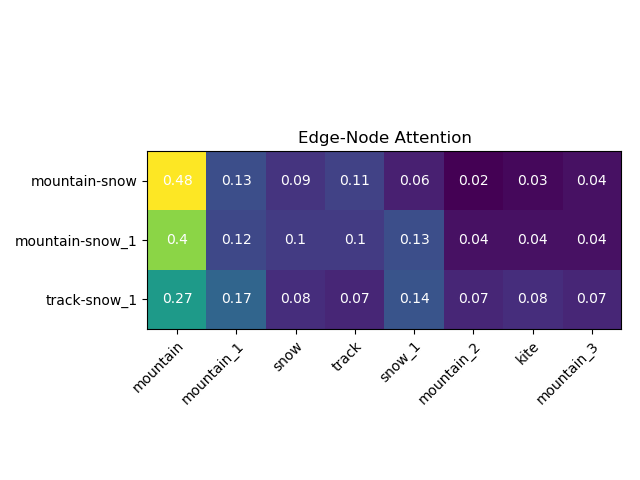}
    \end{subfigure}
    \begin{subfigure}[b]{0.24\linewidth}
        \centering
        \includegraphics[width=\linewidth]{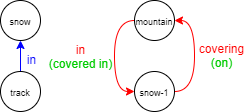}
    \end{subfigure}
    \begin{subfigure}[b]{0.24\linewidth}
        \centering
        \includegraphics[width=\linewidth, height=4cm]{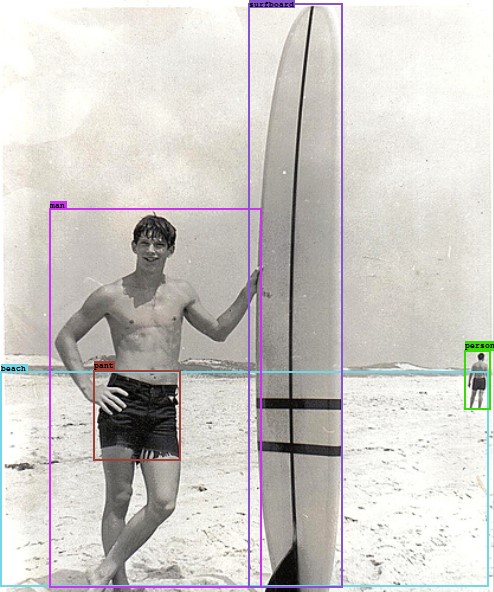}
        \caption{Image}
    \end{subfigure}    
    \begin{subfigure}[b]{0.24\linewidth}
        \centering
        \includegraphics[width=1\linewidth]{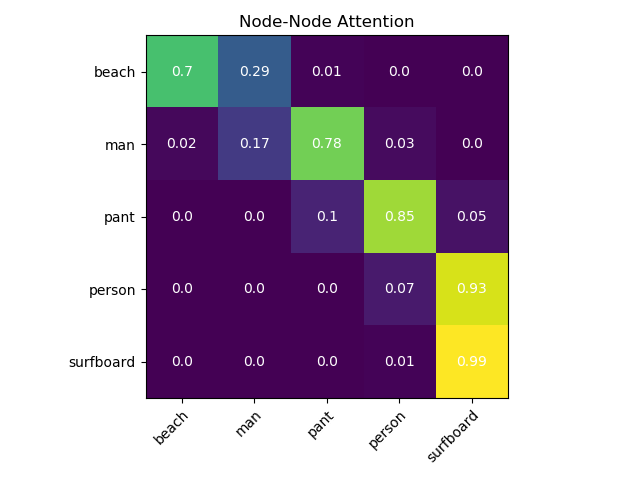}
        \caption{N2N Attention heatmap}
    \end{subfigure}
    \begin{subfigure}[b]{0.24\linewidth}
        \centering
        \includegraphics[width=\linewidth]{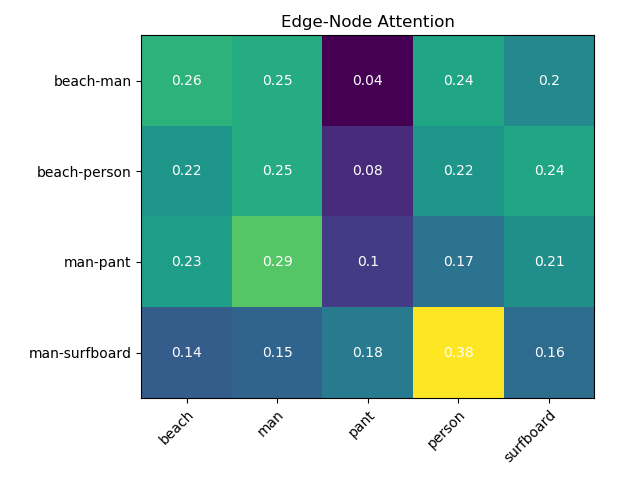}
        \caption{E2N Attention heatmap}
    \end{subfigure}
    \begin{subfigure}[b]{0.24\linewidth}
        \centering
        \includegraphics[width=\linewidth]{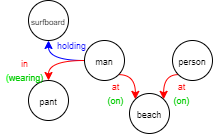}
        \caption{Generated scene graphs}
    \end{subfigure}

\caption{Some negative example outputs from our network with associated attention map and scene graph.}
\label{fig:attention map negative}
\end{figure*}

Fig. \ref{fig:attention map negative}, shows the negative scenario, where network prediction is different from ground truth labels. In most of these cases, it was found that predicted labels are semantically closer to ground truth labels, and from a human perspective, both could be right. For example \textit{man-at-beach} and \textit{man-on-beach} both are grammatically correct.

While exploring various attention heads, we have found an interesting pattern that few attention heads are focusing on the main object in the scene, few on the combination of some objects while others focus on surroundings. Some recent research work\cite{voita2019analyzing} also explores the working patterns for various attention heads.

% \end{thebibliography}

% \end{document}

\end{document}